\documentclass[conference]{IEEEtran}
\IEEEoverridecommandlockouts
\usepackage{cite}
\usepackage{amsmath,amssymb,amsfonts}
\usepackage{graphicx}
\usepackage{textcomp}
\usepackage{xcolor}
\usepackage{hyperref}
\usepackage{multicol}
\usepackage{algorithm,algpseudocode}

\def\BibTeX{{\rm B\kern-.05em{\sc i\kern-.025em b}\kern-.08em
    T\kern-.1667em\lower.7ex\hbox{E}\kern-.125emX}}
\begin{document}

\title{Video Summarisation with Incident and Context Information using Generative AI\\
}

\author{\IEEEauthorblockN{Ulindu De Silva\IEEEauthorrefmark{1},
        Leon Fernando\IEEEauthorrefmark{1},
        Kalinga Bandara\IEEEauthorrefmark{1},
        Rashmika Nawaratne\IEEEauthorrefmark{2}
        } \\
	\IEEEauthorblockA{\IEEEauthorrefmark{1}Department of Electronic and Telecommunication Engineering, University of Moratuwa, Sri Lanka \\
 \IEEEauthorrefmark{2}Research Centre for Data Analytics and Cognition, La Trobe University, Victoria, Australia}
    
}

\maketitle

\begin{abstract}

The proliferation of video content production has led to vast amounts of data, posing substantial challenges in terms of analysis efficiency and resource utilization. Addressing this issue calls for the development of robust video analysis tools. This paper proposes a novel approach leveraging Generative Artificial Intelligence (GenAI) to facilitate streamlined video analysis. Our tool aims to deliver tailored textual summaries of user-defined queries, offering a focused insight amidst extensive video datasets. Unlike conventional frameworks that offer generic summaries or limited action recognition, our method harnesses the power of GenAI to distil relevant information, enhancing analysis precision and efficiency. Employing YOLO-V8 for object detection and Gemini for comprehensive video and text analysis, our solution achieves heightened contextual accuracy. By combining YOLO with Gemini, our approach furnishes textual summaries extracted from extensive CCTV footage, enabling users to swiftly navigate and verify pertinent events without the need for exhaustive manual review. The quantitative evaluation revealed a similarity of 72.8\%, while the qualitative assessment rated an accuracy of 85\%, demonstrating the capability of the proposed method.
\end{abstract}

\begin{IEEEkeywords}
Gemini, Surveillance Video Analysis, Generative Artificial Intelligence, Object Detection
\end{IEEEkeywords}

\section{Introduction}
In recent years, the advancement of generative artificial intelligence has revolutionized various industries, offering unprecedented opportunities for optimization. GenAI  \cite{radford2018improving} \cite{fui2023generative} technologies, leveraging deep learning models, have demonstrated remarkable capabilities in tasks ranging from natural language processing \cite{iorliam2024comparative} to image generation \cite{suryadevara2020generating}. Particularly in industrial settings, where efficiency and precision are paramount, the value of GenAI has become increasingly apparent. This is evidenced by its application in diverse fields such as surveillance \cite{muhammad2019deepres}, manufacturing, and consumer behaviour analysis. The ability of GenAI-based tools to process vast amounts of real-time data and extract actionable insights has positioned them as indispensable assets for businesses striving to stay ahead in today's competitive landscape.

These tools can detect anomalies, defects, unauthorized access, early signs of equipment degradation, bottlenecks and equipment malfunctions in real-time, thereby preventing accidents, thefts, and downtime, and enabling prompt corrective actions ultimately enhancing overall productivity. By monitoring workflows, material flows, and resource utilization, they can suggest process improvements, layout changes, or workflow adjustments to enhance operational efficiency, reduce waste, and maximize throughput. In terms of safety compliance and incident investigation, these prove invaluable by ensuring compliance with safety regulations, monitoring work areas for safety violations, and providing real-time alerts to supervisors. In the event of accidents or incidents, they can analyze video footage to reconstruct events, determine root causes, and facilitate incident investigations, aiding in the development of preventive measures to mitigate future risks.

Over the past decade, hardware advancements leading to increased processing power have facilitated feeding vast amounts of data into complex models. Tech giants like Google, Facebook, Apple, OpenAI, and Microsoft have invested billions in research, yielding super-strong models. We have leveraged Gemini \cite{team2023gemini}, the latest addition, to its fullest potential for video analysis by building on top of it.

This paper proposes a novel video analysis pipeline that leverages GenAI to streamline the analysis process. Our tool goes beyond generic summaries or recognizing a limited set of actions. Instead, it focuses on providing concise textual summaries tailored to user-customizable queries. This approach eliminates irrelevant information, significantly improving the efficiency of video content exploration and retrieval, and catering to the specific needs of industrial applications. The primary contributions of this paper include;
\begin{enumerate}
    \item Specific incident identification through a customizable prompt in videos.
    \item Key incident identification and summarization of videos. 
    \item A novel pipeline for easy-to-use automated video analysis.
    \item A qualitative and quantitative analysis to justify the validity and accuracy of the algorithm.
\end{enumerate}

The rest of the paper is organized as follows: Section \textrm{II} discusses the already existing methods followed by section \textrm{III} with the proposed methodology, and section \textrm{IV} describes the experimental results to demonstrate the effectiveness of the approach. The paper ends with section \textrm{V} followed by section \textrm{VI} with future works. 

\section{Literature Review}

\subsection{General Video Summarization}
General Video Summarization \cite{gunawardena2019interest} aims to provide a high-level overview of a video's content. They offer a quick glimpse but often lack detail and may not capture specific aspects of interest to the viewer. Traditional methods include key-frame extraction  \cite{zhou2018deep}, shot boundary detection (identifying scene changes), and video segmentation  \cite{chen2021simple}. More recently, deep learning models have emerged that can analyze video content and automatically generate summaries, often in the form of short video clips or textual descriptions.

\subsection{Specific Action Recognition}
This approach focuses on identifying predefined actions within videos. \cite{2020mmaction2} While valuable for specific applications, they are limited to a fixed set of actions and may not be adaptable to diverse user queries. Traditional methods rely on handcrafted features and motion analysis techniques to recognize specific activities. Deep learning models, particularly convolutional neural networks (CNNs), have become dominant in this area, achieving high accuracy in recognizing a wide range of actions, such as walking, running, or object manipulation  \cite{kong2022human}.

\subsection{Anomaly Detection}
This technique aims to identify unusual or unexpected events within video data. Traditional methods typically involve statistical analysis of motion patterns or pixel-level changes. Deep learning-based anomaly detection often utilizes auto-encoders  \cite{gong2019memorizing} or one-class classification models to learn the "normal" behaviour within a video. Deviations from this learned behaviour may then be flagged as anomalies \cite{nawaratne2019spatiotemporal}.

\subsection{Object Detection and Tracking}
Identifying and tracking objects within video sequences is a crucial aspect of video analysis. Deep learning models, particularly YOLO  \cite{redmon2016you} and Faster R-CNN  \cite{girshick2015fast}, have become state-of-the-art for object detection, offering high accuracy and real-time processing capabilities  \cite{li2019siamrpn++}. Object tracking algorithms then link detections across frames to follow objects' movements.

\subsection{Keyframe and Skimming-based Video Summarization}
Keyframe-based summaries  \cite{wu2017novel} rely on static content within video samples, such as objects, colours, and other tangible elements, to generate a condensed representation using a selected set of keyframes. However, a notable drawback of keyframe-based approaches is the loss of dynamic information \cite{jadon2020unsupervised} present in the original video. To address this limitation and incorporate dynamic information into the summarization process, skimming-based methods have been developed. Skimming-based summaries  \cite{kumar2019evs} prioritize dynamic content  \cite{truong2007video}, including actions, behaviours, and object movements  \cite{pope1998video}, to create a summary consisting of a combination of dynamic clips.

\subsection{Generative AI for Video Analysis}
The field of video analysis is increasingly embracing GenAI  \cite{radford2018improving}  \cite{huang2021gpt2mvs} techniques to create more informative and user-centric summaries. Auto-regressive video captioning models like Show and Tell  \cite{vinyals2015show} and Transformer-based models  \cite{weng2021event} use a sequence-to-sequence learning framework to generate textual descriptions of video content. Anomaly detection with generative models involves training a model to learn the "normal" behaviour within a video. Deviations from this learned behaviour may then be flagged as anomalies. \cite{huang2022self}
\\
\\
Recent advancements in generative adversarial networks (GANs) \cite{goodfellow2014generative}, particularly conditional GANs (cGANs), have also shown promising results in addressing challenges such as image de-raining in video surveillance. A generative latent-based approach \cite{nawaratne2020generative} has been proposed to tackle real-time road surveillance challenges, arising from adverse weather conditions, lighting levels, motion blur, and low-resolution recordings from CCTV cameras. \cite{hettiarachchi2021rain} These innovative techniques demonstrate significant potential in improving image quality and overcoming practical obstacles in real-time surveillance systems.

\section{Methodology}
\begin{figure*}[h]
    \centering
    \centerline{\includegraphics[width=1\linewidth]{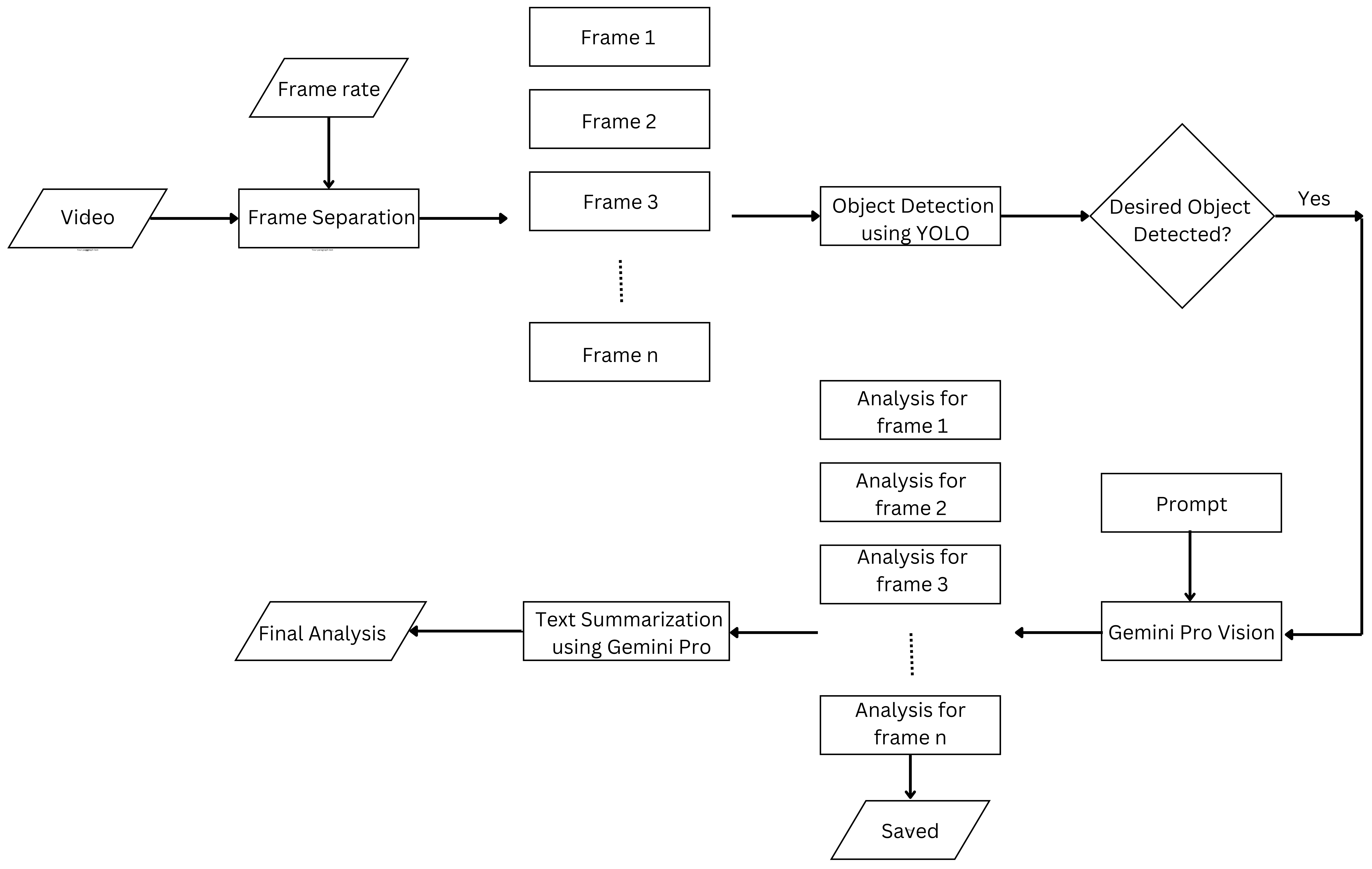}}
    \caption{Proposed Method Overview}
    \label{fig:1}
\end{figure*}

In this section, we propose our novel pipeline for video analysis as illustrated in figure \ref{fig:1}. The video stream first undergoes segmentation into individual frames based on a specified frame rate, tailored to the specific use case. The adjustment of the frame rate allows for flexibility, enabling a higher frame rate for swiftly moving content and a lower frame rate for slower-paced scenarios. Using an unnecessarily higher frame rate produces repeated information at a cost of both time and money. Therefore it is needed to find the appropriate frame separation rate. 

The frames are sequentially fed to a pre-trained Yolo model to detect humans or other desired objects at the runtime. If detected, the segmented frames are fed to the GenAI model, exemplified by the Gemini Pro Vision, where a tailored prompt guides its creative and specific output. The model's responsiveness to contextual nuances is refined through manipulation of the temperature variable, offering flexibility in shaping the generated content. Additionally, the adaptability of the prompt itself allows alignment with the desired output. Based on the expected output, we have to tune the safety settings implied by the harm category, harassment, hate speech, sexual content and dangerous content output of the model. 

\begin{algorithm}
\caption{Key Incident Identification and Summarization}
\begin{algorithmic}[1]
    \State \textbf{Input:} Surveillance video, Prompt for the Gemini Pro
    \State \textbf{Parameters:} Frame rate, the temperature of each Gemini Pro Vision modal and Gemini Pro model, required object to be identified by the YOLO.
    \State \textbf{Output:} Key incident identification and summarization
    \While{video}
        \State Sample the video according to the frame rate.
        \State Input the frame to the YOLO 
        \If{YOLO identifies the specific object required}
            \State Input the frame to the Gemini Pro Vision to describe the given frame
            \State The generated descriptive text is appended to an array with the frame number
        \Else
            \State Continue
        \EndIf
    \EndWhile
    \State Create a paragraph by concatenating the array sentences.
    \State The paragraph is given to the Gemini Pro model for analysis according to the entered prompt.
\end{algorithmic}
\end{algorithm}

The prompts were given in two methods. The following is one sample for figure \ref{fig:3}.
\begin{itemize}
    \item Direct - The prompt asks Gemini Pro Vision to describe if there is an accident happening in the image.
    \item Indirect - The prompt asks Gemini Pro Vision to describe all the happenings in the image. The generated text is then fed to Gemini Pro to see whether an accident was happening in the image.  
\end{itemize}

It was observed that the indirect method produced better results. Gemini Pro Vision was creative in the face of complicated prompts and ended up creating what was not in the image. 

To assess the model's temporal comprehension of the scene, evaluations were conducted using a sequence of images presented frame by frame. Additionally, a collage approach was employed, submitting multiple frames simultaneously. Results indicated that utilizing a sequence of images with an optimized prompt produced superior outcomes.

Attempts to enhance model guidance involved incorporating YOLO  \cite{redmon2016you} crops of the target object as inputs to the Gemini model. While intended to offer guidance, this method did not significantly improve description accuracy. Furthermore, it occasionally led to complications, particularly when the bounding box was not perfectly aligned to scale, resulting in sub-optimal outcomes.

While conducting a frame-wise analysis is a fundamental step, it can lead to redundancy and noise in the results. To address this, the frame-wise output undergoes further processing. It is directed to summarizing using the Gemini Pro model with a prompt, culminating in the extraction of refined and concise final results. This strategic approach enhances the interpretability and utility of the video analysis outcomes. The final result can guide the user to specific frames where he can only check those to find and confirm what he wants to see, instead of using his time checking the entire video manually.  

Standard pipelines can be modified for specific use cases. For instance, analyzing customer interactions with a new retail shop item might involve adding a bounding box layer after frame separation. This guides the model to focus on the region of interest (the new item) for improved analysis.

Sometimes better results are achieved by asking Gemini Pro Vision to describe frame by frame and then ask specific questions from the generated data.  

\section{Experiments and Results}

Quantitative and qualitative experiments were performed on selected videos of the \href{https://huggingface.co/datasets/AlexZigma/msr-vtt/viewer/default/train?p=3&f%5Bvideo_id%5D%5Bmin%5D=9&f%5Bvideo_id%5D%5Bimax%5D=9&f%5Bvideo_id%5D%5Btransform%5D=length&row=398}{MSR-VTT} dataset, and YouTube CCTV videos captured in industrial environments.

\subsection{Dataset}
Microsoft Research-Video to Text, MSR-VTT  \cite{xu2016msr} is a large-scale video benchmark dataset which contains more than 10K web video clips and 200K clip-sentence pairs consisting of 20 categories. We used selected videos from the dataset to experiment with the proposed model and evaluate its performance.

We developed a custom dataset from selected YouTube CCTV videos of supermarkets and roadside footage of accidents and vehicle movements. These videos cover diverse scenarios of human behavior in retail settings and various traffic incidents. Eight volunteers from the Department of Electronic and Telecommunication Engineering at the University of Moratuwa manually generated ground truth summaries of approximately 50 words for each video.


\subsection{Quantitative Analysis}
First, punctuation and stop words are removed from both the generated and predicted sentences. We encode those words using the Glove \cite{pennington2014glove} vectors and if the similarity is higher than 60\% we take them as matched words. Percentage similarity is calculated as follows by comparing the matched words with the ground truth.

\begin{equation}
    Similarity (\%) = \frac{M}{G} \times 100
\end{equation}

where M is the number of matched words and G is the number of ground truth words. Across the dataset, we got a percentage similarity of 72.8.


\begin{figure}[h]
    \centering
    \includegraphics[width=1\linewidth]{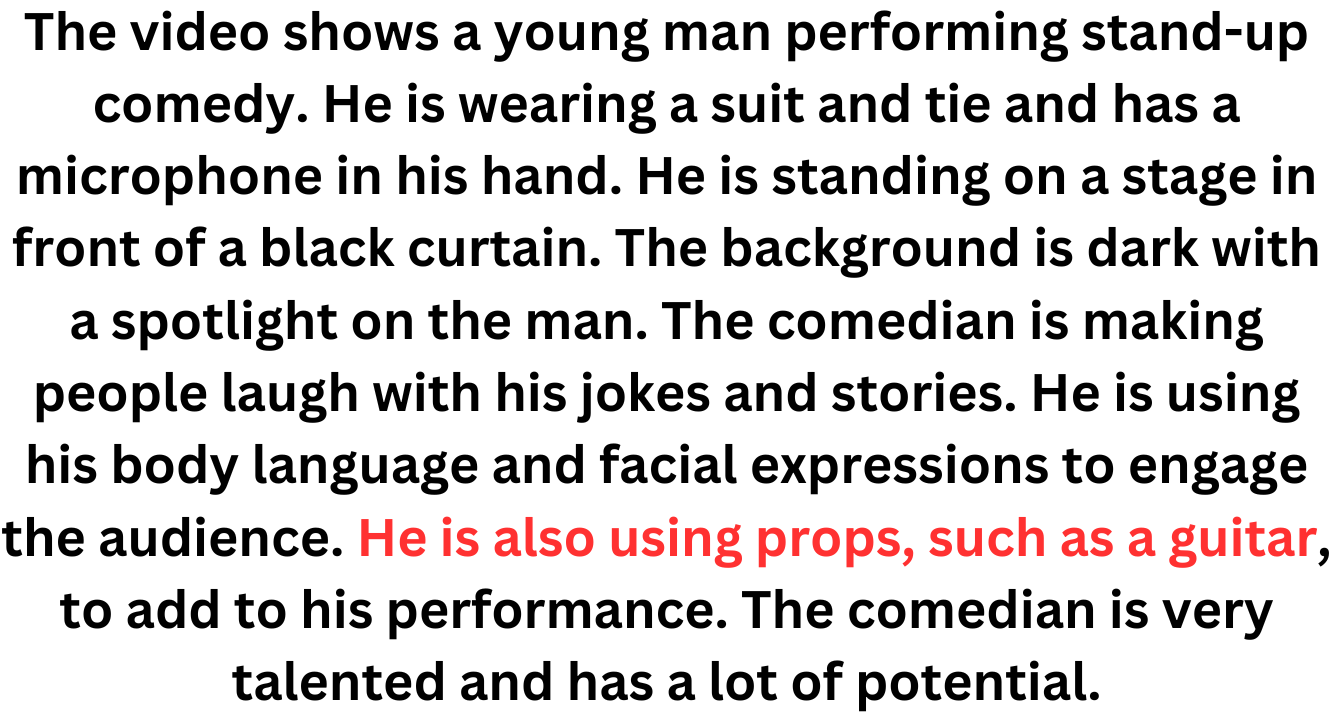}
    \caption{Model output for video extracted from MSR-VTT dataset with video ID: 1360. The model consistently generates high-quality output, although it incorrectly assumes the comedian uses a guitar in the performance when it's merely part of the background. Also, inferring the comedian's talent solely based on this context is speculative, showcasing the model's occasional creativity beyond factual constraints.}
    \label{fig:2}
\end{figure}

Subsequently, we employed ChatGPT to ascertain whether the generated text and the ground truth communicate equivalent meanings by feeding both into the model. ChatGPT affirmed the concordance between the generated text and the ground truth in all the cases.

Regarding time complexity, for Gemini Pro Vision it took an average of 5.5089 s and for summarization using Gemini Pro it took an average of 0.011 s. It is important to note that these times are influenced by the use of online APIs, which means they are subject to variations in internet speed. The processing times can be significantly reduced by deploying an embedded version of the models or utilizing a faster internet connection.

\begin{table}[h]
\caption{Analysis for CCTV traffic video}
\centering
\renewcommand{\arraystretch}{1.5} 
\begin{tabular}{|c|c|p{4.2cm}|}
\hline
\textbf{Timestamp} & \textbf{Frame Number} & \textbf{Information} \\
\hline
00:02 & Frame 2 & Shows the general traffic conditions during the day \\
00:11 & Frame 10 & Shows the traffic situation on the road, with various types of vehicles \\
00:15 & Frame 15 & Shows a motorcycle accident \\
00:17 & Frame 18 & Shows a motorcycle rider who has lost control of his bike \\
00:19 & Frame 20 & Shows a motorcycle rider who has been knocked off his bike by a car. \\
00:22 & Frame 23 & Shows a motorcycle rider who has been injured in a collision with a car. \\
00:28 & Frame 27 & Shows a road with blue and white lines, surrounded by trees and buildings.\\
\hline
\end{tabular}
\label{tab:1}
\end{table}

\begin{figure*}[h]
    \centerline{\includegraphics[width=1\linewidth]{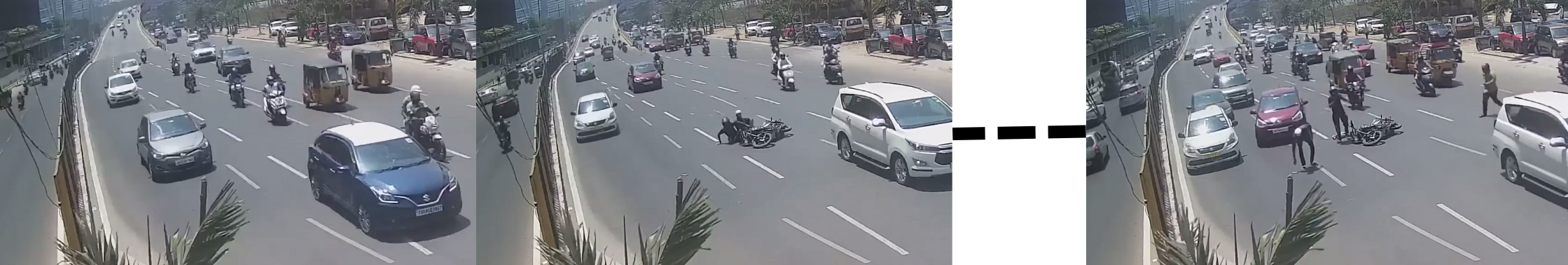}}
    \caption{CCTV traffic video}
    \label{fig:3}
\end{figure*}

\subsection{Qualitative Analysis}
The model exhibits proficiency in accurately identifying key events within the video content. Despite setting the temperature variable to its lowest value, occasional instances such as figure \ref{fig:2} occurred where the model appeared to draw conjectural conclusions. Moreover, the vision model occasionally encounters challenges in accurately discerning colours. 
To conduct a more in-depth examination, we enlisted the assistance of 8 students from the same cohort as in the previous section. They were tasked with evaluating the generated summaries, assigning a score ranging from 1 to 10 based on the model's ability to succinctly summarize the key information present in the video content, with 1 being the idea conveyed from the text is entirely different from the video and 10 being the closest match. On average they rated 8.5 out of 10, which demonstrated the human validation of the summary output generated by our method. 

\subsection{Key incident identification and summarization }

We experimented with how the proposed pipeline identifies key incidents in a video and summarizes the content. First, we show the results observed for a video [figure \ref{fig:3}] from the MSR-VTT dataset with the human-annotated caption. Framewise results for that video is in table \ref{tab:1}.\\ 

\textbf{Gemini Pro Vision Prompt:} Describe the image. 

\textbf{Gemini Pro Prompt:} These are image descriptions of a video. Understand, remove redundant information and give a summary. 

\textbf{Tool generated summary:} The video shows a busy road with various types of vehicles, including cars, motorcycles, auto-rickshaws, and buses. There is heavy traffic, and the vehicles are driving in both directions. The road is lined with trees and buildings. The video also includes footage of a motorcycle accident. 

\textbf{Ground truth:} A motorcycle accident happens on a busy road with many vehicles. \\





This video is of length 29 seconds. We use a frame rate of 1 frame per second. Our method successfully gives out information about the key incidents that happened. The final summarized text generated is rich in detail in contrast with the human-annotated ground truth caption. 
The tool has successfully identified the key incidents with its frame number and timestamp. Finally, the tool has effectively used the frame-wise information to make the summarized text of the whole video.


\begin{table*}[h]
\caption{Accident identification for CCTV traffic video}
\centering
\renewcommand{\arraystretch}{1.5} 
\begin{tabular}{|c|c|p{13cm}|}
\hline
\textbf{Timestamp} & \textbf{Frame Number} & \textbf{Information} \\
\hline
00:13 & Frame 14 &  A motorcycle accident is shown in this frame. A motorcyclist is lying on the ground, injured. A car is stopped next to him. The motorcyclist is wearing a helmet. The car has a damaged front end. The accident happened on a busy road. There are other vehicles in the background.
 \\
00:15 & Frame 15 & A motorcyclist has fallen off his bike and is lying in the middle of the road. A white van is driving in the same direction as the motorcyclist and is approaching him. The van driver swerves to avoid hitting the motorcyclist, but the motorcyclist's bike is hit by a car coming from the opposite direction. The motorcyclist is thrown to the ground and the car that hit him drives away.
 \\
00:17 & Frame 17 & A motorcycle rider has lost control of his bike and fallen to the ground. A white van is driving in the same direction as the motorcycle and is approaching the fallen rider. The van driver swerves to avoid hitting the rider but loses control of the van and crashes into a tree.
\\
00:21 & Frame 20 & This frame depicts a motorcycle accident on a busy road. A motorcyclist has fallen off his bike and is lying in the middle of the road. A man is running towards him to help. There are cars and other vehicles driving around them. The motorcyclist is wearing a helmet, which may have saved his life.
 \\
00:23 & Frame 23 & The image shows a motorcycle rider on a busy road after being knocked off his bike by a car. The rider is lying in the middle of the road, injured. A passerby is helping him. The motorcyclist is helped to his feet and looks dazed. He is then helped to the side of the road as cars continue to drive by.
 \\
00:24 & Frame 24 & The picture shows the aftermath of a motorcycle accident. A motorcycle rider has crashed into the back of a car. The motorcyclist is lying on the ground, injured. The car driver is standing next to his car, looking at the motorcyclist. There is a traffic jam behind the accident.
 \\
\hline
\end{tabular}
\label{tab:2}
\end{table*}

\subsection{Specific incident identification}
Using a prompt the pipeline can be modified to identify specific incidents that take place in a video. We used a CCTV video of a road accident as in figure \ref{fig:3} and asked whether an accident occurred in the video.

Qualitative analysis shows the proposed pipeline can accurately filter the frames containing the incidents of road accidents. Further, the tool has given a brief description of the accidents in those frames. [Table: \ref{tab:2}] \\

\textbf{Gemini Pro Vision Prompt:} Describe the image.  

\textbf{Gemini Pro Prompt:} These are frame-wise descriptions of a video. Understand and describe the frames containing accidents.

\section{Conclusion}
This paper proposed a novel pipeline for video analysis. The main benefit of the proposed method is that we can reduce the time and cost taken for traditional video analysis where the analyser has to evaluate the entire video. When compared with novel anomaly detection or action recognition models, our approach can generate more descriptions dynamically without selecting from a discrete set. To evaluate the feasibility and the effectiveness of the proposed method we have conducted a series of tests on the MSR-VTT dataset. The experiment results show that the model is capable of accurately displaying what is happening in the video. The qualitative analysis further shows that the model is capable of key instance recognition with a customizable prompt.

\section{Future Works}
In the proposed method, we use Gemini Pro vision and Gemini Pro as the video analysis models. But in an Industrial setting using a cloud-based model might not always be commercially efficient due to both API calls costing money and a lot of time being taken to upload the image and download the output. So we intend to experiment with a local server, or running on an onboard computer. To enhance the model's performance for a specific industrial application, fine-tuning it with specific images related to that industrial setting is necessary. Further, we plan to make a comprehensive surveillance dataset, make ground truth and evaluate the performance of the dataset. 

\section{Acknowledgment}
The authorship team would like to acknowledge the vision, support and guidance of the IEEE Industrial Electronics Society in conducting the Generative AI Hackathon under the leadership of Daswin De Silva and Lakshitha Gunasekara.

\bibliographystyle{ieeetr}
\bibliography{sample}

\begin{thebibliography}{10}

\bibitem{radford2018improving}
A.~Radford, K.~Narasimhan, T.~Salimans, I.~Sutskever, {\em et~al.}, ``Improving language understanding by generative pre-training,'' 2018.

\bibitem{fui2023generative}
F.~Fui-Hoon~Nah, R.~Zheng, J.~Cai, K.~Siau, and L.~Chen, ``Generative ai and chatgpt: Applications, challenges, and ai-human collaboration,'' 2023.

\bibitem{iorliam2024comparative}
A.~Iorliam and J.~A. Ingio, ``A comparative analysis of generative artificial intelligence tools for natural language processing,'' {\em Journal of Computing Theories and Applications}, vol.~2, no.~1, pp.~91--105, 2024.

\bibitem{suryadevara2020generating}
C.~K. Suryadevara, ``Generating free images with openai’s generative models,'' {\em International Journal of Innovations in Engineering Research and Technology}, vol.~7, no.~3, pp.~49--56, 2020.

\bibitem{muhammad2019deepres}
K.~Muhammad, T.~Hussain, J.~Del~Ser, V.~Palade, and V.~H.~C. De~Albuquerque, ``Deepres: A deep learning-based video summarization strategy for resource-constrained industrial surveillance scenarios,'' {\em IEEE Transactions on Industrial Informatics}, vol.~16, no.~9, pp.~5938--5947, 2019.

\bibitem{team2023gemini}
G.~Team, R.~Anil, S.~Borgeaud, Y.~Wu, J.-B. Alayrac, J.~Yu, R.~Soricut, J.~Schalkwyk, A.~M. Dai, A.~Hauth, {\em et~al.}, ``Gemini: a family of highly capable multimodal models,'' {\em arXiv preprint arXiv:2312.11805}, 2023.

\bibitem{gunawardena2019interest}
P.~Gunawardena, H.~Sudarshana, O.~Amila, R.~Nawaratne, D.~Alahakoon, A.~S. Perera, and C.~Chitraranjan, ``Interest-oriented video summarization with keyframe extraction,'' in {\em 2019 19th International Conference on Advances in ICT for Emerging Regions (ICTer)}, vol.~250, pp.~1--8, IEEE, 2019.

\bibitem{zhou2018deep}
K.~Zhou, Y.~Qiao, and T.~Xiang, ``Deep reinforcement learning for unsupervised video summarization with diversity-representativeness reward,'' in {\em Proceedings of the AAAI conference on artificial intelligence}, vol.~32, 2018.

\bibitem{chen2021simple}
W.~Chen, X.~Du, F.~Yang, L.~Beyer, X.~Zhai, T.-Y. Lin, H.~Chen, J.~Li, X.~Song, Z.~Wang, {\em et~al.}, ``A simple single-scale vision transformer for object localization and instance segmentation,'' {\em arXiv preprint arXiv:2112.09747}, 2021.

\bibitem{2020mmaction2}
M.~Contributors, ``Openmmlab's next generation video understanding toolbox and benchmark.'' \url{https://github.com/open-mmlab/mmaction2}, 2020.

\bibitem{kong2022human}
Y.~Kong and Y.~Fu, ``Human action recognition and prediction: A survey,'' {\em International Journal of Computer Vision}, vol.~130, no.~5, pp.~1366--1401, 2022.

\bibitem{gong2019memorizing}
D.~Gong, L.~Liu, V.~Le, B.~Saha, M.~R. Mansour, S.~Venkatesh, and A.~v.~d. Hengel, ``Memorizing normality to detect anomaly: Memory-augmented deep autoencoder for unsupervised anomaly detection,'' in {\em Proceedings of the IEEE/CVF international conference on computer vision}, pp.~1705--1714, 2019.

\bibitem{nawaratne2019spatiotemporal}
R.~Nawaratne, D.~Alahakoon, D.~De~Silva, and X.~Yu, ``Spatiotemporal anomaly detection using deep learning for real-time video surveillance,'' {\em IEEE Transactions on Industrial Informatics}, vol.~16, no.~1, pp.~393--402, 2019.

\bibitem{redmon2016you}
J.~Redmon, S.~Divvala, R.~Girshick, and A.~Farhadi, ``You only look once: Unified, real-time object detection,'' in {\em Proceedings of the IEEE conference on computer vision and pattern recognition}, pp.~779--788, 2016.

\bibitem{girshick2015fast}
R.~Girshick, ``Fast r-cnn,'' in {\em Proceedings of the IEEE international conference on computer vision}, pp.~1440--1448, 2015.

\bibitem{li2019siamrpn++}
B.~Li, W.~Wu, Q.~Wang, F.~Zhang, J.~Xing, and J.~Yan, ``Siamrpn++: Evolution of siamese visual tracking with very deep networks,'' in {\em Proceedings of the IEEE/CVF conference on computer vision and pattern recognition}, pp.~4282--4291, 2019.

\bibitem{wu2017novel}
J.~Wu, S.-h. Zhong, J.~Jiang, and Y.~Yang, ``A novel clustering method for static video summarization,'' {\em Multimedia Tools and Applications}, vol.~76, pp.~9625--9641, 2017.

\bibitem{jadon2020unsupervised}
S.~Jadon and M.~Jasim, ``Unsupervised video summarization framework using keyframe extraction and video skimming,'' in {\em 2020 IEEE 5th International Conference on computing communication and automation (ICCCA)}, pp.~140--145, IEEE, 2020.

\bibitem{kumar2019evs}
K.~Kumar, ``Evs-dk: Event video skimming using deep keyframe,'' {\em Journal of Visual Communication and Image Representation}, vol.~58, pp.~345--352, 2019.

\bibitem{truong2007video}
B.~T. Truong and S.~Venkatesh, ``Video abstraction: A systematic review and classification,'' {\em ACM transactions on multimedia computing, communications, and applications (TOMM)}, vol.~3, no.~1, pp.~3--es, 2007.

\bibitem{pope1998video}
A.~Pope, R.~Kumar, H.~Sawhney, and C.~Wan, ``Video abstraction: Summarizing video content for retrieval and visualization,'' in {\em Conference Record of Thirty-Second Asilomar Conference on Signals, Systems and Computers (Cat. No. 98CH36284)}, vol.~1, pp.~915--919, IEEE, 1998.

\bibitem{huang2021gpt2mvs}
J.-H. Huang, L.~Murn, M.~Mrak, and M.~Worring, ``Gpt2mvs: Generative pre-trained transformer-2 for multi-modal video summarization,'' in {\em Proceedings of the 2021 International Conference on Multimedia Retrieval}, pp.~580--589, 2021.

\bibitem{vinyals2015show}
O.~Vinyals, A.~Toshev, S.~Bengio, and D.~Erhan, ``Show and tell: A neural image caption generator,'' in {\em Proceedings of the IEEE conference on computer vision and pattern recognition}, pp.~3156--3164, 2015.

\bibitem{weng2021event}
W.~Weng, Y.~Zhang, and Z.~Xiong, ``Event-based video reconstruction using transformer,'' in {\em Proceedings of the IEEE/CVF International Conference on Computer Vision}, pp.~2563--2572, 2021.

\bibitem{huang2022self}
C.~Huang, J.~Wen, Y.~Xu, Q.~Jiang, J.~Yang, Y.~Wang, and D.~Zhang, ``Self-supervised attentive generative adversarial networks for video anomaly detection,'' {\em IEEE transactions on neural networks and learning systems}, 2022.

\bibitem{goodfellow2014generative}
I.~Goodfellow, J.~Pouget-Abadie, M.~Mirza, B.~Xu, D.~Warde-Farley, S.~Ozair, A.~Courville, and Y.~Bengio, ``Generative adversarial nets,'' {\em Advances in neural information processing systems}, vol.~27, 2014.

\bibitem{nawaratne2020generative}
R.~Nawaratne, S.~Kahawala, S.~Nguyen, and D.~De~Silva, ``A generative latent space approach for real-time road surveillance in smart cities,'' {\em IEEE Transactions on Industrial Informatics}, vol.~17, no.~7, pp.~4872--4881, 2020.

\bibitem{hettiarachchi2021rain}
P.~Hettiarachchi, R.~Nawaratne, D.~Alahakoon, D.~De~Silva, and N.~Chilamkurti, ``Rain streak removal for single images using conditional generative adversarial networks,'' {\em Applied Sciences}, vol.~11, no.~5, p.~2214, 2021.

\bibitem{xu2016msr}
J.~Xu, T.~Mei, T.~Yao, and Y.~Rui, ``Msr-vtt: A large video description dataset for bridging video and language,'' in {\em Proceedings of the IEEE conference on computer vision and pattern recognition}, pp.~5288--5296, 2016.

\bibitem{pennington2014glove}
J.~Pennington, R.~Socher, and C.~D. Manning, ``Glove: Global vectors for word representation,'' in {\em Proceedings of the 2014 conference on empirical methods in natural language processing (EMNLP)}, pp.~1532--1543, 2014.

\end{thebibliography}





\end{document}